\documentclass[letterpaper]{article} 
\usepackage[preprint]{aaai2027}
\usepackage[hyphens]{url}  
\usepackage{graphicx} 
\usepackage{natbib}  
\usepackage{caption} 
\usepackage{algorithm}
\usepackage{algorithmic}
\usepackage{booktabs}

\usepackage{amsfonts}
\usepackage{amsmath}
\usepackage{amssymb}
\usepackage{nicefrac}
\usepackage{array}
\usepackage{multirow}
\usepackage{tabularx}

\newcommand{\method}{\textsc{GeoSurge}}

\newcommand{\privdataset}{\textsc{GeoSurge-Bench}}
\newcommand{\crlmdataset}{\textsc{CRLM}}
\newcommand{\medagent}{\textsc{3DMedAgent}}
\newcommand{\deeptumor}{\textsc{DeepTumorVQA}}
\newcommand{\mewm}{\textsc{MeWM}}

\newcommand{\points}{\mathbf{P}}
\newcommand{\facts}{\mathbf{F}}

\newcommand{\ltokens}{\widetilde{\mathbf{Z}}}

\newcommand{\cmark}{\ensuremath{\checkmark}}
\newcolumntype{L}[1]{>{\raggedright\arraybackslash}p{#1}}
\newcolumntype{Y}{>{\raggedright\arraybackslash}X}
\newcolumntype{C}[1]{>{\centering\arraybackslash}p{#1}}

\title{\method{}: Geometry-Grounded Multimodal Reasoning for Auditable Liver Resection Planning}
\makeatletter
\newcommand{\markJointCorrespondence}{\aaai@corrmultitrue}
\makeatother

\author{%
  Wenjie Deng\textsuperscript{\rm 1}\equalcontrib,
  Zijian Tu\textsuperscript{\rm 1}\equalcontrib,
  Xueyin Zhou\textsuperscript{\rm 1}\equalcontrib,
  Huayi Wang\textsuperscript{\rm 1},\\
  Zhuoyun Du\textsuperscript{\rm 1},
  Mingyu Chen\textsuperscript{\rm 2}\markJointCorrespondence\corresponding,
  Haochao Ying\textsuperscript{\rm 1}\corresponding,
  Jian Wu\textsuperscript{\rm 1}
}
\affiliations{%
  \textsuperscript{\rm 1}Zhejiang University, Hangzhou, China\\
  \textsuperscript{\rm 2}Sir Run Run Shaw Hospital, Zhejiang University School of Medicine, Hangzhou, China
}

\begin{document}

\maketitle

\begin{abstract}
Surgical resection planning centers on generating executable resection surfaces from anatomical evidence and operative constraints. Medical multimodal foundation models enable anatomical understanding, and segmentation models predict resection regions, but neither translates these capabilities into explicitly constrained, executable resection surfaces. We present \method{}, an anatomy-grounded framework that combines mask-derived metrics with structure-aware point-cloud tokens to generate executable, editable resection surfaces.
To make resection geometry learnable and editable, \method{} maps actions to parametric geometry through a surface-action grammar. 
Using this grammar, anatomy-oriented and tumor-centered templates generate candidates, which are filtered by hard constraints and semantic deduplication.
Training supervision is then derived from candidate–expert geometric agreement and independently recomputed feasibility metrics. At inference, the reasoner uses patient evidence and surgical requirements to select or revise an action, which an independent geometry engine subsequently materializes and verifies.
\method{} achieves the best overall scores on DeepTumour Liver and CRLM-QA, improving the constraint satisfaction rate from 73\% to 100\% on CRLM and from 56\% to 99\% on \privdataset{}.
\end{abstract}

\section{Introduction}


Preoperative resection simulation enables iterative exploration of alternative resection plans before irreversible tissue transection. Across liver, kidney, and lung surgery, virtual planning has demonstrated the potential to refine surgical decision-making and improve selected perioperative outcomes~\cite{mise2018virtual,shirk2019effect,he2025three,lu2023printing}.
Conventional planners construct explicit surfaces and compute volumetric or vascular effects, but rely on surgeon-drawn traces and iterative deformation~\cite{lang2005virtual,palomar2016surface}. Recent medical foundation models offer anatomy-conditioned inference. Together, these advances provide complementary capabilities for anatomy-aware reasoning, explicit surface manipulation, and quantitative assessment of resection plans. 
Building on these capabilities, multimodal large language models have the potential to integrate clinical knowledge with patient-specific evidence to iteratively simulate alternative resection strategies and predict consequences, reducing clinicians’ planning burden while supporting safer, outcome-aware decisions and a higher likelihood of success.

\begin{figure}[t]
\centering
\includegraphics[width=\columnwidth]{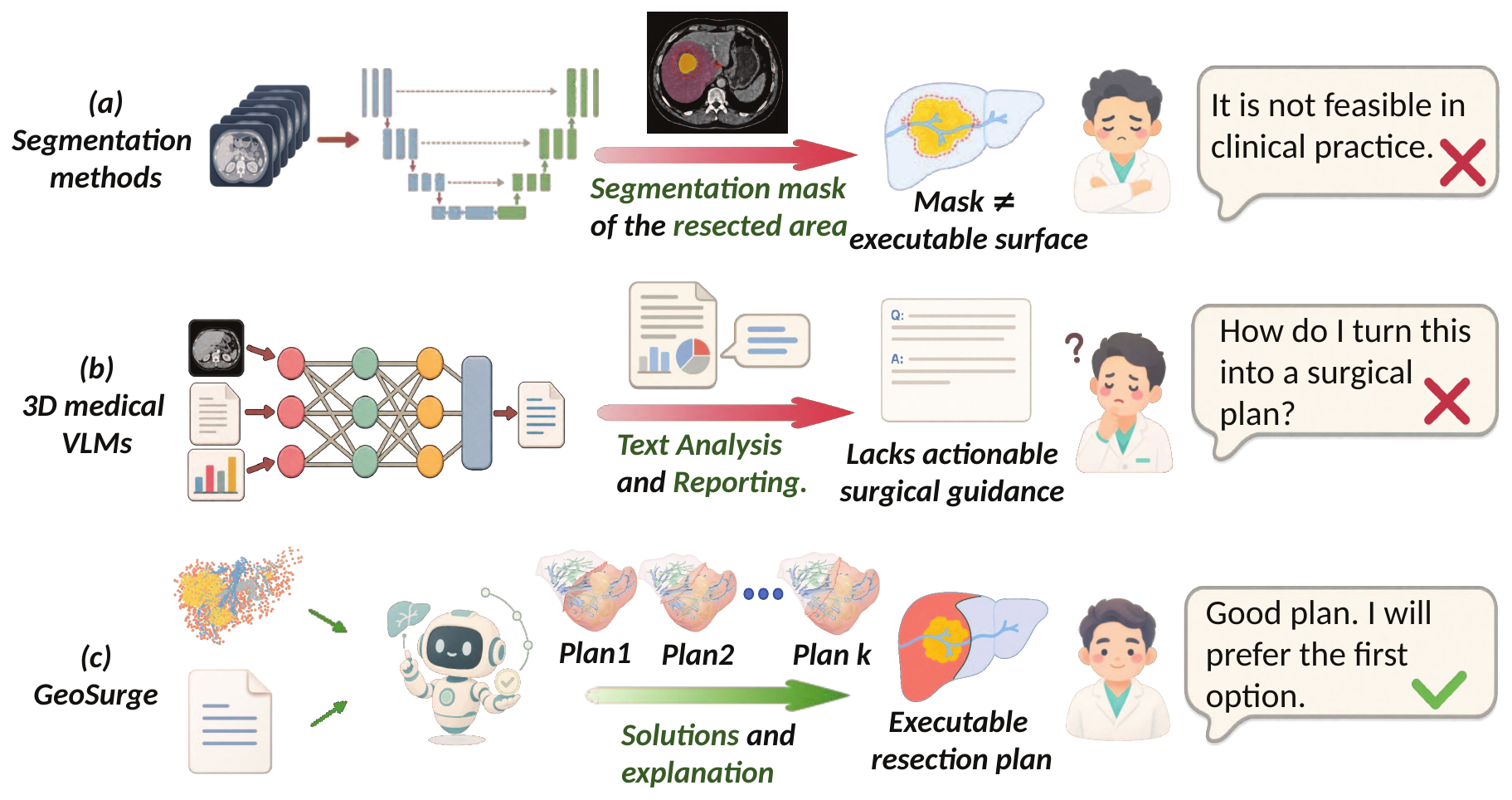}
\caption{Existing methods and \method{} for executable liver-resection planning.
(a) \textbf{Segmentation methods} predict masks without explicit anatomical constraints.
(b) \textbf{3D medical VLMs} reason about anatomy but lack actionable geometry.
(c) \textbf{\method{}} generates anatomically constrained, executable surfaces to support surgical planning.}
\label{fig:teaser}
\end{figure}
Recent advances in medical image segmentation and multimodal foundation models have improved region delineation and anatomical reasoning. Segmentation models learn to predict expert-defined resection regions from voxel- or point-level annotations~\cite{rakshit2025lizard}. As shown in Fig.~\ref{fig:teaser}(a), voxel-wise supervision reduces the transection surface to an unconstrained mask boundary, encoding the expert resection policy in model weights rather than as explicit geometric and safety constraints. 
Multimodal large language models have shown strong performance in report generation,
VQA, localization, segmentation, and tool-based analysis, demonstrating anatomical and spatial
reasoning capabilities~\cite{wu2023radfm,bai2024m3d,wang2026medagent}.
However, as shown in Fig.~\ref{fig:teaser}(b), their interfaces
and training corpora encode neither parameterized transection surfaces nor their post-resection effects. Consequently, patient-specific resection surface planning has not yet been formulated as a learnable, anatomy-conditioned geometric decision problem.

Executable surface planning poses two coupled challenges. First, the central representation question is whether anatomical evidence can retain both metric fidelity and fine-grained 3D structure under the compression required for language-model reasoning. Volumetric MLLMs efficiently capture global 3D context, but resampling, patchification, and spatial pooling do not guarantee that sub-patch boundaries or thin vascular structures remain distinguishable. The resulting tokens also do not determine physical scale without voxel spacing~\cite{bai2024m3d,trinh2026spatialmed}. Tool-derived measurements preserve exact physical quantities, but their scalar outputs do not encode the local tumor--vessel geometry required for surface placement~\cite{wang2026medagent}. Second, resection planning involves a high-dimensional output space and noisy supervision. Continuous transection surfaces contain numerous geometric degrees of freedom, making them difficult to represent compactly, interpret semantically, and learn effectively. Moreover, existing datasets provide only masks of completed resections, in which anatomical requirements are entangled with surgeon preferences and procedural variations, preventing them from providing clean supervision for model training~\cite{simpson2024crlm}.

\begin{figure*}[t]
\centering
\includegraphics[width=\textwidth]{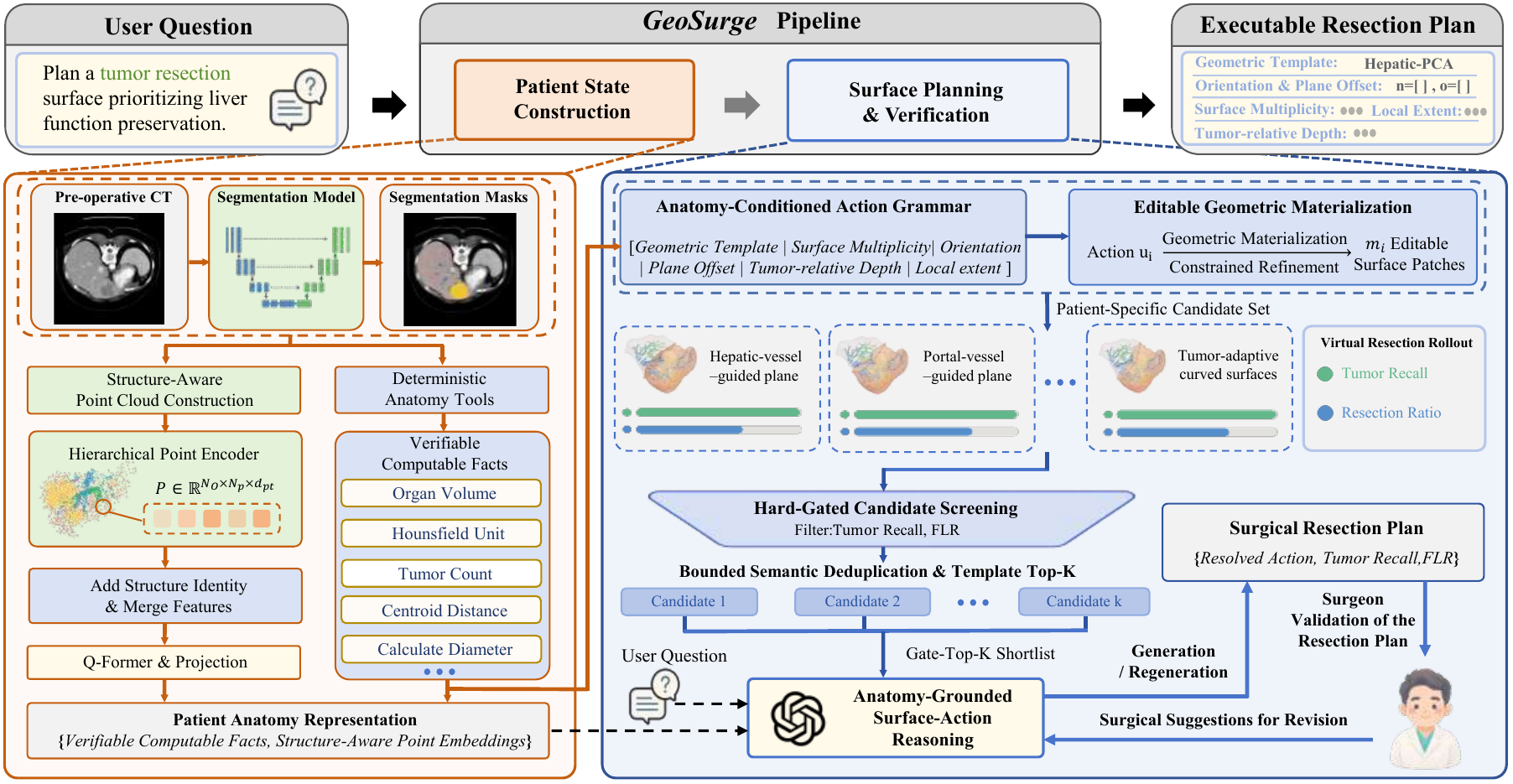}
\caption{\textbf{\method{} pipeline.}
(1) \textbf{Patient representation.}
CT-derived masks produce verifiable facts and structure-aware point embeddings.
(2) \textbf{Candidate compilation.}
An action grammar generates editable surfaces, followed by virtual rollout, hard gating, deduplication, and Top-$K$ selection.
(3) \textbf{Reasoning and verification.}
Conditioned on the request and patient evidence, the reasoner selects or revises an action, which an independent geometry engine materializes and verifies.}
\label{fig:pipeline}
\end{figure*}

Motivated by these observations, we present \method{}, a framework that converts preoperative anatomy and surgical requirements into executable and editable resection surfaces (Fig.~\ref{fig:pipeline}). 
Structure-aware point embeddings and verifiable computable facts jointly serve as model inputs. For the former, predicted masks are converted into structure-tagged point clouds, which a hierarchical point encoder and query bridge map into compact tokens. The executable surface-action grammar represents each candidate plan by its template, surface count, orientation, offset, depth, and extent. These fields deterministically map to one or more editable parametric surfaces. \method{} generates candidate actions from anatomy-oriented and tumor-centered surface templates, then removes infeasible and redundant candidates before reasoning. During dataset construction, we compare candidate plans with expert resections and independently recompute their anatomical effects. These results train the reasoner to select a candidate, complete missing action fields, identify infeasible plans, and revise a plan for a stated preference. Training teaches the action reasoner to perform these four tasks. At inference, the reasoner resolves an action, after which an independent geometry engine materializes the corresponding surface, recomputes its post-resection effects for verification, and supports further parametric revisions to align the plan with the user’s intended resection strategy. Our contributions are summarized as follows:
\begin{itemize}
    \item We formulate resection-surface planning, for the first time, as a learnable geometric decision-making problem over executable actions grounded in anatomical evidence and operative constraints.
    \item We propose \method{}, integrating recomputable anatomical
    facts and structure-aware point clouds with constrained
    actions, candidate-conditioned reasoning, and deterministic verification.
    
    \item Evaluations across two reasoning benchmarks and two surface-planning cohorts demonstrate consistent improvements in geometric understanding and planning.
    
\end{itemize}

\section{Related Work}

\paragraph{3D Medical Reasoning.}
Recent 3D medical reasoning spans volumetric vision--language representation, task-specific
evaluation, and tool-supported evidence analysis.
RadFM, M3D-LaMed, Merlin, and CT-CHAT extend medical vision--language modeling to volumetric
images~\cite{wu2023radfm,bai2024m3d,blankemeier2024merlin,hamamci2024ctrate}.
DeepTumorVQA evaluates tumor-centric recognition, measurement, and higher-order reasoning, while
\medagent{} combines tool orchestration and structured memory for evidence-driven CT
analysis~\cite{chen2025deeptumor,wang2026medagent}. Together, these studies advance volumetric
reasoning and evidence organization, but their outputs remain centered on interpretation,
measurement, and textual reporting rather than clinical planning artifacts. They therefore do not
connect patient-specific evidence to executable geometry that clinicians can inspect and edit, with
quantities that can be independently recomputed.

\paragraph{Geometry-Aware Liver Planning.}
Liver resection planning spans classical surgical modeling, learned region prediction, and
foundation segmentation.
Computer-assisted liver planning models vasculature, virtual resection, operative risk, and future
liver remnant~\cite{selle2002vasculature,meinzer2002planning,lang2005virtual,palomar2016surface}.
Learned geometric baselines span point-cloud segmentation with RandLA-Net and its liver-resection
adaptation LIZARD-RandLA-Net~\cite{hu2020randlanet,rakshit2025lizard}, together with volumetric
segmentation using nnU-Net and STU-Net~\cite{isensee2021nnunet,huang2023stunet}. Segmentation
foundation models further provide patient-specific anatomical masks, while action-conditioned models
such as \mewm{} study post-treatment states~\cite{wasserthal2023totalsegmentator,he2025vista3d,yang2025mewm}.
These methods supply anatomical masks, resection regions, and outcome estimates, but cover only
parts of the clinical surface decision. Segmentation and resection-zone masks delineate target tissue
without defining an explicit, clinically executable transection surface; classical planners construct
patient-specific surfaces under predefined objectives rather than case-conditioned requests and
surgical preferences. More broadly, unconstrained prediction in a continuous, high-dimensional
surface space neither guarantees geometric validity nor provides a natural interface for inspection,
editing, and recomputation.

\section{Method}
\label{sec:method}

\subsection{Problem Formulation}

\method{} avoids direct high-dimensional surface regression by compiling patient anatomy into a
bounded, editable, and verifiable surface-action space over which an LLM can reason. Given a
preoperative CT $X$, predicted anatomical masks $\mathcal M$, and a request $q$, the task is to return an
explicit transection surface $S^\star$ whose anatomical effects can be recomputed independently.
Each legal action
$\mathbf u_i\in\mathcal U(\mathcal M)$ is materialized as geometry $S_i$ and paired with
recomputable outcomes $\mathbf e_i$, yielding
\begin{equation}
\mathcal C_0(\mathcal M)=
\left\{c_i=(\mathbf u_i,S_i,\mathbf e_i):
\mathbf u_i\in\mathcal U(\mathcal M)\right\}_{i=1}^{N}.
\label{eq:candidate_bank}
\end{equation}
Feasibility screening and redundancy removal produce a bounded set
$\mathcal C_K\subseteq\mathcal C_0$. The reasoner receives a request, auditable facts $\facts$,
structure-aware tokens $\ltokens$, and candidate cards
$\mathcal Z_K=\{\mathbf z_i\}_{c_i\in\mathcal C_K}$. It predicts a grammar-level decision $a^\star$
that selects or revises typed action fields before they are resolved into executable geometry.
\begin{equation}
\begin{aligned}
a^\star&=\arg\max_{a\in\Omega(q,\mathcal Z_K)}
p_\theta(a\mid q,\facts,\ltokens,\mathcal Z_K),\\
\mathbf u^\star&=\operatorname{Resolve}(a^\star,\mathcal C_K),
\qquad S^\star=G(\mathbf u^\star).
\end{aligned}
\label{eq:candidate_conditional}
\end{equation}
Here $\Omega$ includes complete-action decisions and typed parameter changes. Surface
materialization and geometric verification remain independent of the model.

\subsection{Patient Anatomy Representation}

Exact metric evidence and local 3D organization provide complementary views of patient anatomy. We
therefore combine a canonical computable-fact schema with structure-aware point embeddings.

\paragraph{Verifiable computable facts.}
A deterministic toolkit maps the inputs to a shared fact schema
$\facts=f_{\mathrm{fact}}(X,\mathcal M)$, providing exact and auditable anatomical measurements and
relations. Conditioned on the task, the reasoner attends to the relevant evidence in $\facts$, while
the expert resection strategy remains hidden. Full schema definitions and task-specific derivations
are provided in the supplementary appendix.

\paragraph{Structure-aware point embeddings.}
We stack liver, tumor, portal vein, and hepatic vein samples as
$\points\in\mathbb R^{N_{\mathcal O}\times N_p\times5}$, where
$\mathcal O=\{\mathrm{liv},\mathrm{tum},\mathrm{pv},\mathrm{hv}\}$,
$N_{\mathcal O}=|\mathcal O|=4$, and $N_p$ is the number of points
sampled from each structure. For each structure $o\in\mathcal O$,
$\points_o\in\mathbb R^{N_p\times5}$ denotes its structure-specific
point set. Each point is represented by
$[\hat{x},\hat{y},\hat{z},\hat{h},\hat{s}]$, comprising normalized
liver-centered coordinates, windowed HU intensity, and a normalized
liver-scale descriptor. All structures are expressed in a shared
coordinate frame while being encoded separately to preserve their
anatomical identities. A hierarchical point
encoder~\cite{qi2017pointnetpp} extracts structure-specific features. Following the Q-Former
design~\cite{li2023blip2}, a lightweight query bridge produces
\begin{equation}
\ltokens=Q_\omega\!\left(
\mathop{\mathrm{Concat}}_{o\in\mathcal O}
\left[E_\psi(\points_o)+e_{\mathrm{type}}(o)\right]\right).
\label{eq:anatomy_tokenizer}
\end{equation}
Thus $\facts$ anchors exact quantities, while $\ltokens$ preserves local and cross-structure
geometry. Alignment objectives are provided in the supplementary appendix.

\subsection{Executable Surface Action Compilation}

\paragraph{Anatomy-conditioned action grammar.}
Continuous surfaces form a high-dimensional, non-identifiable search space for an LLM. We instead
compile complementary geometries into a bounded grammar with anatomy-oriented plane and
tumor-centered local-surface templates. Each typed action
$\mathbf u_i=(f_i,m_i,\nu_i,\delta_i,d_i,\ell_i)$ specifies its geometric template $f_i$, surface
multiplicity $m_i$, orientation $\nu_i$, plane offset
$\delta_i$, tumor-relative depth $d_i$, and local extent or shape $\ell_i$. Only the fields valid for
the selected template are instantiated. Local actions follow the case-specific tumor-cluster count and
grouping, whereas anatomy-oriented actions follow patient-specific axes and corridors. Enumerating
legal anatomy-derived values produces the raw bank in Eq.~\eqref{eq:candidate_bank} without expert
resection masks or test-time labels. Template fields are listed in the appendix.

\paragraph{Editable geometric materialization.}
Discrete actions are suitable for reasoning but too coarse for executable transection and subsequent
clinical editing, so each is lifted into a higher-dimensional parametric surface. The constructor
instantiates $\mathbf u_i$ as $m_i$ editable surface patches, with multiplicity determined by the
selected template. 
Each patch is defined over a finite spatial support in a local frame
$(\mathbf o,\mathbf n_S,\mathbf e_1,\mathbf e_2)$, using the following B\'ezier form:
\begin{equation}
S(s,t)=\mathbf o+\alpha(s)\mathbf e_1+\beta(t)\mathbf e_2
+\mathbf n_S\,\mathbf b(s)^\top H\mathbf b(t).
\label{eq:bezier_surface}
\end{equation}
Here $\alpha,\beta$ map normalized coordinates to the local support, $\mathbf b$ is a Bernstein
basis, and $H$ supplies local degrees of freedom for nonplanar boundaries; its control points remain available for manual secondary editing, as in interactive surface planning~\cite{palomar2016surface}.
Starting from the anatomy-conditioned action, constrained refinement targets 5~mm tumor
clearance while regularizing displacement, curvature, and boundary motion. Each materialized
surface then undergoes a virtual resection rollout, where the induced resection region and
post-resection anatomical effects are recomputed from the masks, including tumor recall,
resection ratio, vessel involvement, detachment, and stability.

\paragraph{Hard-gated candidate screening.}
Geometric validity alone does not ensure surgical feasibility, so noncompensatory constraints must be
enforced before learned reasoning.
Applying a candidate surface $S_i$ to the preoperative liver and tumor masks defines its induced
liver resection region $R_i\subseteq M_{\mathrm{liv}}$. We compute tumor recall
$r_i=|R_i\cap M_{\mathrm{tum}}|/|M_{\mathrm{tum}}|$, resection ratio
$\rho_i=|R_i|/|M_{\mathrm{liv}}|$, and FLR fraction $\eta_i=1-\rho_i$. Thus FLR is recomputed from
the tissue retained by the candidate surface, rather than read from an expert FLR annotation. With thresholds
$\boldsymbol\tau_g=(\tau_{\mathrm{rec}},\tau_{\mathrm{flr}})$, the feasible set is:
\begin{equation}
\mathcal C_1=\left\{c_i\in\mathcal C_0:
r_i\geq\tau_{\mathrm{rec}},\ 
\eta_i\geq\tau_{\mathrm{flr}}
\right\}.
\label{eq:hard_gate}
\end{equation}
The gate rejects any violation before ranking, and no scalar score can restore a rejected candidate.


\paragraph{Bounded semantic deduplication.}
Anatomical enumeration produces many near-duplicates, whereas the reasoner requires a compact yet
template-diverse context.
Within the feasible set, semantic deduplication collapses candidates that share a geometric template,
multiplicity, and semantic axis and have near-identical recomputable endpoints. Template-Top-$K$ then
preserves template coverage before filling the remaining slots by the anatomy-only order. We denote
this hard-gate--deduplication--Top-$K$ pipeline as Gate-Top-$K$. Shortlisting remains deterministic and model-hidden, and rejected candidates are never restored by subsequent scoring.
Full pseudocode, endpoint matching rules, and tie-breakers are provided in the supplementary appendix.

\subsection{Anatomy-Grounded Surface Action Reasoning}

\method{} makes high-dimensional surface planning accessible to an LLM through the shared action
grammar instantiated by $\mathcal C_K$. The reasoner learns how template-specific fields relate to
patient anatomy, enabling it to complete parameters, compare actions, diagnose failures, and revise
a surface parameterization under a changed intent rather than acting only as a candidate ranker.

\paragraph{Action-grammar supervision.}
One case-level candidate bank supports four action-level tasks. Surface selection compares complete
actions against patient evidence; parameter completion recovers a missing active grammar field;
safety diagnosis identifies the principal violated constraint; and preference adjustment revises the
parameterization under a changed intent. Together, these tasks train one reasoner to compare,
complete, diagnose, and edit executable actions. Construction details appear in the appendix.

\paragraph{Instruction-Conditioned Revision.}
Given an operator-specified parameter change, \method{} maps
the instruction to a bounded surface-action edit and rechecks anatomical constraints. This enables
plan refinement without redrawing a 3D mask or rerunning the full pipeline. 
Although we did not conduct a clinician study, we designed an idealized one-edit correction experiment to estimate the potential benefit of limited clinician intervention.
\paragraph{Separated construction and model input.}
We derive CRLM-QA targets offline from public expert resection annotations~\cite{simpson2024crlm} and the
recomputed outcomes of the compiled bank. Expert-reference agreement supervises surface selection
and parameter completion, while recomputed feasibility outcomes supervise safety diagnosis and
preference adjustment. A separate serializer then constructs every model input exclusively
as $(q,\facts,\ltokens,\mathcal Z_K)$; each card in $\mathcal Z_K$ contains only its identifier and
typed grammar parameters. The offline construction record is never loaded by the reasoner. Exact task
rules, accepted-answer handling, visible fields, and split-wise counts are provided in the appendix.

\paragraph{Staged optimization.}
Stage~1 trains the point encoder by point--text alignment and reconstruction of the predicted masks. Stage~2 freezes it
and Qwen2.5-7B~\cite{qwen2024qwen25}, updating only the Q-Former bridge and projections, PointAlign
projector, and LoRA~\cite{hu2022lora}. It jointly optimizes anatomical QA and the four action tasks
without rationale supervision.

\paragraph{Independent geometric verification.}
The verifier recomputes tumor recall, FLR, vessel involvement, detachability, and stability without
altering the decision or feeding results to the reasoner. Logged shortlist, action, surface, seed,
threshold, and diagnostic records preserve provenance.
\begin{table*}[htbp]
\centering
{\small
\setlength{\tabcolsep}{3.0pt}
\renewcommand{\arraystretch}{1.10}
\begin{tabular*}{\textwidth}{@{\extracolsep{\fill}}l|c|ccccc|ccccc@{}}
\toprule
\multicolumn{1}{c|}{\multirow{2}{*}{\textbf{Model}}}
& \multirow{2}{*}{\textbf{Param.}}
& \multicolumn{5}{c|}{\textbf{DeepTumour Liver}}
& \multicolumn{5}{c}{\textbf{CRLM-QA}} \\
\cline{3-7}\cline{8-12}
& & Quant. & Lesion & Spatial & Tissue & Mean
& Select. & Complete & Safety & Pref. & Mean \\
\midrule

Rand
& --
& 0.25 & 0.39 & 0.29 & 0.40 & 0.34
& 0.41 & 0.69 & 0.17 & 0.17 & 0.36 \\

\mbox{Qwen3-VL}
& \mbox{30B-A3B}
& \underline{0.32} & 0.51 & 0.24 & 0.36 & 0.31
& 0.38 & \underline{0.75} & 0.21 & 0.28 & 0.40 \\

\mbox{Qwen2.5-VL}
& 7B
& 0.26 & 0.24 & 0.41 & 0.34 & 0.31
& 0.35 & 0.71 & 0.09 & 0.16 & 0.33 \\

MedGemma
& 27B
& 0.26 & 0.45 & 0.57 & 0.30 & 0.40
& 0.36 & \underline{0.75} & 0.24 & 0.27 & 0.41 \\

\mbox{HuatuoGPT-Vision}
& 34B
& 0.28 & 0.65 & 0.32 & 0.35 & 0.40
& 0.42 & 0.72 & \underline{0.29} & \underline{0.31} & \underline{0.44} \\

\mbox{M3D-LaMed-Phi-3}
& 4B
& 0.24 & 0.78 & 0.22 & 0.28 & 0.38
& 0.42 & 0.71 & 0.23 & 0.17 & 0.38 \\

\mbox{RadFM}
& 14B
& 0.25 & \textbf{0.84} & 0.07 & 0.25 & 0.35
& \underline{0.43} & 0.68 & 0.18 & 0.19 & 0.37 \\

\medagent{}
& \mbox{30B}
& \textbf{0.80} & \underline{0.81} & \underline{0.66}
& \textbf{0.75} & \underline{0.75}
& 0.39 & 0.74 & 0.15 & 0.25 & 0.38 \\

\midrule

\textbf{\method{}}
& 7B
& \textbf{0.80} & \textbf{0.84} & \textbf{0.78}
& \underline{0.69} & \textbf{0.78}
& \textbf{0.64} & \textbf{0.87} & \textbf{0.59}
& \textbf{0.67} & \textbf{0.69} \\

\bottomrule
\end{tabular*}
}
\caption{Comparison results on the DeepTumour Liver and CRLM-QA benchmarks. The ``Rand'' row reports
random-guess accuracy for each task. Best results are bold and second-best results are underlined.}
\label{tab:liverqa_structure_understanding}
\end{table*}

\section{Experiments}

We evaluate \method{} along four axes: anatomical understanding and action reasoning; executable,
constraint-satisfying geometry and external transfer; component contributions; and sensitivity to
shortlist size, candidate scale, and missing or corrupted computable facts.
\subsection{Experimental Setup}

\paragraph{Datasets.}
We evaluate on two public datasets and an independent hospital cohort.
\begin{itemize}
    \item \textbf{DeepTumour Liver} is an anatomical-QA dataset derived from
    \deeptumor{}~\cite{chen2025deeptumor}. It covers quantification,
    lesion recognition, spatial relations, and tissue context, and serves
    as the anatomy benchmark.
    \item \textbf{\crlmdataset{}}~\cite{simpson2024crlm} provides liver, tumor, vascular, and
    future-remnant segmentations from which expert resection references are derived. It supports
    action-level planning QA and executable-surface evaluation.
    \item \textbf{\privdataset{}} is an external cohort of $68$ quality-controlled,
    de-identified pre/postoperative CT pairs from a separate hospital, under institutional ethics
    approval with informed consent waived. It evaluates external transfer. After isotropic resampling
    and deformable registration, its reference is the preoperative liver minus the registered postoperative remnant.
\end{itemize}
We use four-way exact-match accuracy for DeepTumour Liver and accepted-option accuracy for the
CRLM-QA action tasks, counting invalid responses as incorrect. Materialized surfaces on \crlmdataset{}
and \privdataset{} are evaluated by tumor recall, Global Dice, absolute FLR error, and CCSR:
\[
\mathrm{CCSR}
=
\frac{1}{N}
\sum_{i=1}^{N}
\mathbf{1}\!\bigl[
  \textstyle\bigwedge_{r\in\mathcal{R}} a_i^r
\bigr],
\]
where $\mathcal R=\{\mathrm{surf},\mathrm{tum},\mathrm{mar},\mathrm{flr}\}$ denotes surface validity,
tumor coverage sufficiency, margin adequacy, and FLR sufficiency, and $a_i^r$ indicates whether case
$i$ satisfies requirement $r$. CCSR is noncompensatory, whereas Global Dice quantifies geometric
agreement~\cite{reinke2024metrics,antonelli2022msd}. Dataset statistics, splits, metrics, tests, and isolation controls appear in the appendix.

\paragraph{Baselines.}
We compare \method{} with baselines spanning generic multimodal reasoning, medical VLMs, native
3D medical systems, and task-specific geometric predictors. For anatomical QA, we include random
prediction; Qwen3-VL-30B-A3B and Qwen2.5-VL-7B~\cite{bai2025qwen3vl,bai2025qwen25vl};
MedGemma-27B-IT, HuatuoGPT-Vision-34B, and RadFM-14B~\cite{sellergren2025medgemma,
chen2024huatuogptvision,wu2023radfm};
and M3D-LaMed-Phi-3-4B and \medagent{}~\cite{bai2024m3d,wang2026medagent}. General-purpose and 2D medical MLLMs receive 10 uniformly sampled slices from each CT resampled to 60 axial slices, whereas M3D-LaMed and RadFM use the full volume and \medagent{} follows its native evidence pipeline. All methods share the same questions, options, and scorers, making this a system-level rather than strictly input-controlled comparison. For surface planning, geometry baselines include RandLA-Net,
LIZARD-RandLA-Net, nnU-Net, and STU-Net~\cite{hu2020randlanet,rakshit2025lizard,
isensee2021nnunet,huang2023stunet}. Under the matched Executable Surface Action Compilation (ESAC) protocol,
\method{} and all ESAC baselines use the same VISTA3D-predicted masks,
computable facts, candidate cards, and action definitions for every case.
Each selected record is materialized as an editable B\'ezier surface before
scoring. Table~\ref{tab:surface_ablation_robustness} includes direct 32D decoding as a component ablation.

\begin{table*}[!t]

\centering

{\small

\setlength{\tabcolsep}{2.2pt}

\renewcommand{\arraystretch}{1.02}

\begin{tabular*}{\textwidth}{@{\extracolsep{\fill}}lcccccccc@{}}
\toprule
\multicolumn{1}{c}{\multirow{2}{*}{\textbf{Method}}}
& \multicolumn{4}{c}{\textbf{CRLM}}
& \multicolumn{4}{c}{\textbf{\privdataset{}}} \\
\cmidrule(lr){2-5}\cmidrule(lr){6-9}
& G. Dice $\uparrow$ & FLR Err. $\downarrow$ & CCSR $\uparrow$ & Tumor Rec. $\uparrow$
& G. Dice $\uparrow$ & FLR Err. $\downarrow$ & CCSR $\uparrow$ & Tumor Rec. $\uparrow$ \\
\midrule
LIZARD-RandLA-Net
& 0.57 & 0.55 & 0.00 & \underline{0.98}
& 0.42 & 0.61 & 0.00 & 0.78 \\

nnU-Net
& 0.58 & 0.37 & 0.00 & 0.90
& 0.42 & 0.45 & 0.01 & 0.70 \\

STU-Net
& \textbf{0.66} & \textbf{0.17} & 0.07 & 0.90
& 0.37 & 0.27 & 0.01 & 0.60 \\

\midrule

Qwen3-VL-30B-A3B+ESAC
& 0.58 & 0.23 & \underline{0.73} & \textbf{1.00}
& 0.45 & \underline{0.17} & \underline{0.56} & \underline{0.99} \\

Qwen2.5-VL-7B+ESAC
& 0.29 & 0.36 & 0.50 & 0.57
& 0.11 & 0.29 & 0.19 & 0.25 \\

MedGemma-27B+ESAC
& \underline{0.64} & 0.25 & 0.47 & \textbf{1.00}
& \textbf{0.54} & 0.20 & 0.47 & \textbf{1.00} \\

HuatuoGPT-Vision-34B+ESAC
& 0.10 & 0.41 & 0.07 & 0.13
& 0.07 & 0.29 & 0.02 & 0.12 \\

M3D-LaMed-Phi-3-4B+ESAC
& 0.00 & 0.44 & 0.00 & 0.00
& 0.00 & 0.32 & 0.00 & 0.00 \\

\medagent{}-30B+ESAC
& 0.60 & \underline{0.22} & \underline{0.73} & \textbf{1.00}
& 0.49 & \textbf{0.16} & 0.46 & \textbf{1.00} \\

\midrule

\textbf{\method{}}
& \underline{0.64} & \textbf{0.17} & \textbf{1.00} & \textbf{1.00}
& \underline{0.53} & 0.18 & \textbf{0.99} & \textbf{1.00} \\

\method{}+One-Edit Guidance$^\dagger$
& 0.67 & 0.15 & 1.00 & 1.00
& 0.54 & 0.17 & 0.99 & 1.00 \\

\bottomrule
\end{tabular*}

    }
\caption{Surface-planning results on the CRLM and \privdataset{} benchmarks.
Upper rows are geometry baselines; lower rows use ESAC to construct and screen a shared action set and materialize editable B\'ezier surfaces.
Both \method{} rows use $K=16$/cap $0.80$ (post hoc).
$^\dagger$Hidden-reference one-edit oracle; rankings exclude this row.
Bold and underline mark best and second-best results.}
\label{tab:experiment2_surface_planning_framework}
\end{table*}

\begin{figure*}[!t]
\centering
\includegraphics[width=\textwidth]{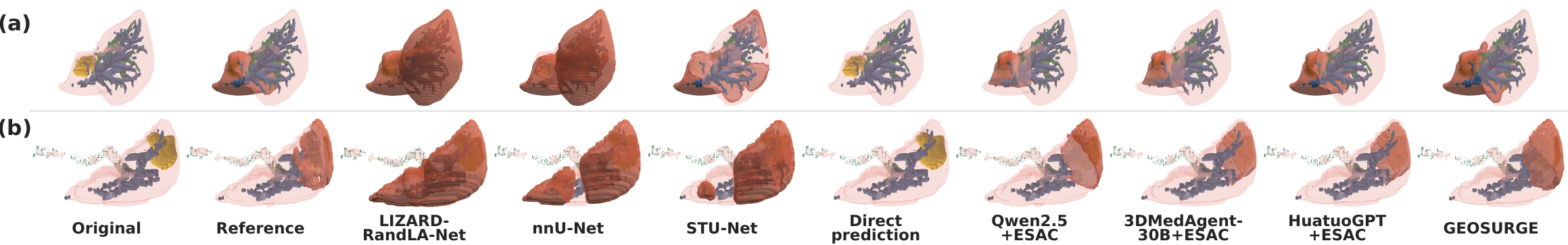}
\caption{Qualitative planning comparison on (a) CRLM case 1103 and (b) an anonymized GeoSURGE-Bench case.
Columns show the original anatomy, reference resection, LIZARD-RandLA-Net, nnU-Net, STU-Net, direct 32D prediction, Qwen2.5-VL+ESAC, \medagent{}-30B+ESAC, HuatuoGPT-Vision+ESAC, and \method{}.}
\label{fig:qualitative_planning_comparison}
\end{figure*}

\begin{table*}[!t]
\centering

{\small
\setlength{\tabcolsep}{1.10pt}
\renewcommand{\arraystretch}{0.98}

\begin{tabular*}{\textwidth}{
@{\extracolsep{\fill}}
>{\raggedright\arraybackslash}p{2.15cm}
cccccccccccc
@{}
}
\toprule

\multicolumn{1}{c}{\multirow{2}{*}{\textbf{Variant}}}
& \multicolumn{4}{c}{\textbf{Components}}
& \multicolumn{4}{c}{\textbf{CRLM}}
& \multicolumn{4}{c}{\textbf{GeoSURGE-Bench}} \\

\cmidrule(lr){2-5}
\cmidrule(lr){6-9}
\cmidrule(lr){10-13}

& FT
& 3D
& Cand.
& Gate
& G. Dice $\uparrow$
& FLR Err. $\downarrow$
& CCSR $\uparrow$
& Tumor Rec. $\uparrow$
& G. Dice $\uparrow$
& FLR Err. $\downarrow$
& CCSR $\uparrow$
& Tumor Rec. $\uparrow$ \\

\midrule

Qwen2.5-VL-7B
&  &  &  &
& 0.00 & 0.44 & 0.00 & 0.00
& 0.00 & 0.32 & 0.00 & 0.00 \\

\midrule

Direct prediction
& \cmark & \cmark &  &
& \underline{0.61} & 0.39 & 0.27 & \textbf{1.00}
& 0.46 & 0.58 & 0.16 & \textbf{1.00} \\

Random Top-$K$
& \cmark & \cmark & \cmark &
& 0.60 & 0.32 & \underline{0.56} & \textbf{1.00}
& 0.48 & 0.41 & 0.54 & \textbf{1.00} \\

Metric only
&  &  & \cmark & \cmark
& 0.55 & 0.21 & \textbf{1.00} & \textbf{1.00}
& 0.47 & \textbf{0.17} & \underline{0.96} & \textbf{1.00} \\

Facts only
& \cmark &  & \cmark & \cmark
& \underline{0.61} & \underline{0.19} & \textbf{1.00} & \textbf{1.00}
& \underline{0.52} & \textbf{0.17} & \textbf{0.99} & \textbf{1.00} \\

\textbf{Full \method{}}
& \cmark & \cmark & \cmark & \cmark
& \textbf{0.64} & \textbf{0.17} & \textbf{1.00} & \textbf{1.00}
& \textbf{0.53} & \underline{0.18} & \textbf{0.99} & \textbf{1.00} \\

\bottomrule
\end{tabular*}
}

\caption{Ablation results on CRLM and GeoSURGE-Bench; terms are defined in text;
bold/underline mark best/second-best.}
\label{tab:surface_ablation_robustness}

\end{table*}

\paragraph{Implementation.}
For both training and testing, a frozen pretrained VISTA3D
model~\cite{he2025vista3d} predicts the liver, tumor, portal vein, and
hepatic vein masks. The same masks define the $N_p=1{,}024$ point samples,
computable facts, and candidate cards. Expert resection references are used
only for offline supervision and evaluation. \method{} uses
Qwen2.5-7B-Instruct~\cite{qwen2024qwen25} as a frozen language backbone. Stage~1 trains the point
encoder with point--text alignment and mask reconstruction; Stage~2 freezes the point encoder and
Qwen, then updates the Q-Former bridge, projections, PointAlign projector, and LoRA adapters while
jointly sampling anatomical QA and the four action tasks. We optimize trainable parameters with
AdamW in bfloat16 and evaluate with deterministic point sampling, greedy decoding, and the same
geometry compiler for materialized plans. Training and inference use one NVIDIA A100-SXM4-80GB GPU. The appendix reports all
hyperparameters, random seeds, and module states.

\subsection{Executable Surface Planning}

Table~\ref{tab:liverqa_structure_understanding} evaluates anatomical and action reasoning on
DeepTumour Liver and CRLM-QA. Table~\ref{tab:experiment2_surface_planning_framework} evaluates
executable planning on CRLM and \privdataset{}.

Executable planning depends on both the geometric output interface and reasoning backbone
(Table~\ref{tab:experiment2_surface_planning_framework}). Table~\ref{tab:liverqa_structure_understanding}
exposes a representation bottleneck rather than a model-scaling advantage. The 2D VLMs preserve local appearance or lesion semantics from sparse slices but do not encode continuous cross-slice geometry or physical spacing. M3D-LaMed and RadFM introduce volumetric encoders, yet fixed-token compression favors global recognition over fine spatial structure, yielding lesion accuracies of 0.78 and 0.84 but spatial accuracies of only 0.22 and 0.07. \medagent{} improves anatomical QA by externalizing masks, measurements, and selected regions into structured memory, although such evidence is not trained for candidate-level counterfactual reasoning. The consistently lower safety and preference scores on CRLM-QA further show that anatomical recognition alone does not support the interactions required between geometry, candidate parameters, and surgical constraints.

Beyond anatomical understanding, Table~\ref{tab:experiment2_surface_planning_framework} compares region prediction with constrained action selection for producing constraint-satisfying surfaces. For the geometry-prediction baselines, LIZARD-RandLA-Net uses local point aggregation, while nnU-Net and STU-Net learn multiscale voxel representations. Their overlap objectives treat the transection surface as an incidental mask boundary. Thus, STU-Net achieves 0.66 Dice and 0.17 FLR error on CRLM but only 0.07 CCSR, showing that region accuracy does not ensure executable feasibility. ESAC alleviates this output-space mismatch by restricting prediction to materializable, hard-gated candidates, but performance remains backbone-dependent. MedGemma leads in geometric overlap, whereas Qwen3-VL and \medagent{} lead in CRLM CCSR. Medical alignment preserves tumor semantics, but sparse or compressed visual tokens weaken the encoding of metric tumor--vessel geometry, while image--text objectives do not train joint interactions among planning constraints.

Full \method{} achieves the strongest executable planning performance across both cohorts.
All multimodal reasoners use the same hard-gated candidates, so the gain comes from better
patient-specific action selection rather than feasibility screening. Aligning 3D evidence and
recomputable facts with typed actions yields CCSR values of 1.00 and 0.99 with complete tumor recall.
We also report a hidden-reference one-edit oracle on the frozen \method{} output. It selects the best
constraint-preserving predefined edit and estimates local correctability rather than clinician-guided
performance.

\subsection{Qualitative Analysis}

Figure~\ref{fig:qualitative_planning_comparison} visualizes how the compared methods place resection
geometry in the same patient-specific anatomy. Across the two cases, the mask- and point-based
baselines produce volumetric boundaries whose topology and extent can differ markedly from the
clinician-defined reference, while direct 32D prediction can fail to materialize a visible resection
surface. The ESAC-based reasoners and \method{} produce explicit surfaces but select different
orientations and extents, illustrating the patient-specific effect of action selection.

\subsection{Architecture Analysis}

Table~\ref{tab:surface_ablation_robustness} disentangles candidate-constrained
prediction, deterministic Gate-Top-$K$ construction, non-LLM metric selection,
and structure-aware 3D evidence. Compared with direct 32D parameter prediction,
introducing the patient-specific candidate set through Random Top-$K$ raises
CCSR from 0.27 to 0.56 on CRLM and from 0.16 to 0.54 on \privdataset{}, while
substantially reducing FLR error. Dice remains comparable, indicating that the
action interface primarily improves constraint satisfaction rather than overlap.

The metric-only variant isolates deterministic non-LLM selection over the shared
Gate-Top-$K$ shortlist. Relative to Random Top-$K$, it raises CCSR from 0.56 to
1.00 on CRLM and from 0.54 to 0.96 on \privdataset{}, while reducing FLR error
from 0.32 to 0.21 and from 0.41 to 0.17, respectively. However, Dice decreases
from 0.60 to 0.55 and from 0.48 to 0.47, showing that explicit metrics strongly
enforce feasibility but do not fully recover expert-aligned geometry.

Replacing the metric selector with the facts-only reasoner improves Dice from
0.55 to 0.61 on CRLM and from 0.47 to 0.52 on \privdataset{}, while preserving
or improving CCSR and FLR error. Adding point tokens further raises Dice to 0.64
and 0.53; FLR error decreases from 0.19 to 0.17 on CRLM but increases slightly
from 0.17 to 0.18 on \privdataset{}, with CCSR unchanged at 1.00 and 0.99.
Tumor recall remains 1.00 throughout. Overall, deterministic selection establishes
executability, whereas learned facts and 3D evidence refine patient-specific geometry.
\subsection{Sensitivity and Robustness Analysis}

\begin{figure}[t]
\centering
\includegraphics[width=\columnwidth]{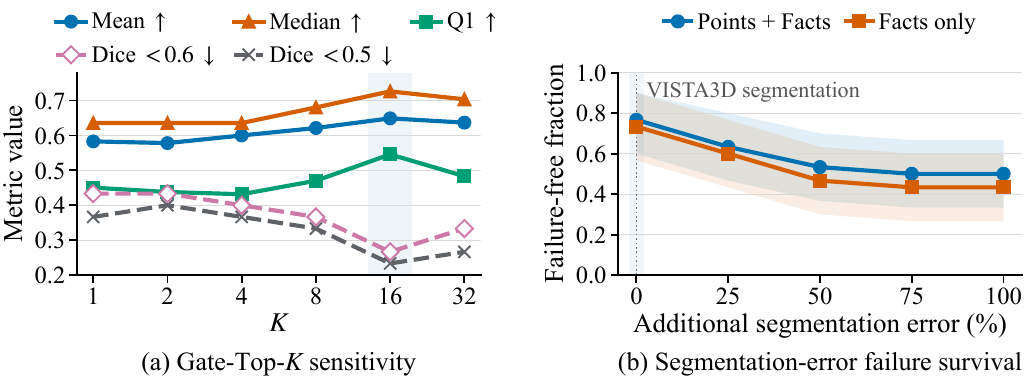}
\caption{Post-hoc sensitivity on 30 CRLM cases.
(a) Dice statistics across Gate-Top-$K$ shortlist sizes (Q1: first quartile).
(b) Failure-free fraction beyond observed segmentation error (coverage-aware
Dice $\geq 0.35$ at every preceding dose; unadjusted 95\% paired-case
bootstrap CIs).}
\label{fig:sensitivity_robustness}
\end{figure}

\paragraph{Gate-Top-$K$.}
Figure~\ref{fig:sensitivity_robustness}(a) shows a non-monotonic
coverage--distraction trade-off. Expanding the feasible shortlist to $K=16$
improves both the center and lower tail of the Dice distribution, suggesting
that narrow gates prune useful alternatives; the decline at $K=32$ indicates
added selection ambiguity. We therefore select $K=16$ for geometric quality
and case-level stability. CCSR instead peaks at $K=4$ (0.967 vs.\ 0.800),
showing that smaller gates favor constraint satisfaction and that $K=16$ is a
deliberate balance rather than a uniformly optimal setting.

\paragraph{Additional segmentation error.}
Starting from the observed-error anchor, both variants lose failure-free cases
as additional corruption accumulates, but Points+Facts remains modestly higher
(Fig.~\ref{fig:sensitivity_robustness}(b)). All 30 cases remain in the
denominator, including those already below the threshold at the anchor. At
$\tau=0.35$, normalized post-anchor survival AUC is 0.575 versus 0.521
(paired $\Delta=0.054$, 95\% CI [$-0.017$, 0.150]). This direction is
consistent with structure-aware evidence and deliberate training-time noise
buffering unreliable facts.

\section{Conclusion}

\method{} moves multimodal surgical planning from anatomical interpretation to executable,
patient-specific geometry. It makes high-dimensional transection planning tractable through an
anatomy-constrained action space of editable surfaces grounded in recomputable evidence, while
separating learned decisions from deterministic assessment. The framework remains dependent on
segmentation and is evaluated at a limited scale. We identify extending the framework across
institutions, anatomies, and procedures as the next step toward general geometry-grounded planning.

\renewcommand{\bibfont}{\small}
\bibliography{references}


\end{document}